\documentclass{article} 

\usepackage{arxiv}

\usepackage[utf8]{inputenc} 
\usepackage[T1]{fontenc}    
\usepackage{booktabs}       
\usepackage{amsfonts}       
\usepackage{nicefrac}       
\usepackage{lipsum}

\usepackage{graphicx}
\usepackage{amssymb}
\usepackage{url}
\usepackage{microtype}

\title{Deep Neural Networks for the Correction of Mie Scattering in
  Fourier-Transformed Infrared Spectra of Biological Samples}

\author{Arne P. Raulf \And Joshua Butke \And Lukas Menzen \And Claus K{\"u}pper \And Frederik Gro{\ss}erueschkamp \And Klaus Gerwert \And Axel Mosig \AND \\
Center for Protein Diagnostics\\Ruhr-University Bochum\\ Gesundheitscampus 4\\44801 Bochum, Germany\\\texttt{axel.mosig@bph.rub.de}}

\begin{document}

\maketitle

\begin{abstract}
  Infrared spectra obtained from cell or tissue specimen have commonly
  been observed to involve a significant degree of (resonant) Mie
  scattering, which often overshadows biochemically relevant spectral
  information by a non-linear, non-additive spectral component in
  Fourier transformed infrared (FTIR) spectroscopic
  measurements. Correspondingly, many successful machine learning
  approaches for FTIR spectra have relied on preprocessing procedures
  that computationally remove the scattering components from an
  infrared spectrum.

  We propose an approach to approximate this complex preprocessing
  function using deep neural networks. As we demonstrate, the
  resulting model is not just several orders of magnitudes faster,
  which is important for real-time clinical applications, but also
  generalizes strongly across different tissue types. Furthermore,
  our proposed method overcomes the trade-off between computation time
  and the corrected spectrum being biased towards an artificial
  reference spectrum.
\end{abstract}

\section{Introduction}
Fourier transform infrared (FTIR) spectroscopic imaging of
biological samples provides pixel spectra at high spatial resolution
which carry a highly informative fingerprint of the biochemical status
of the sample. FTIR microscopy thus has been applied successfully in
characterizing the disease state of tissue samples of different types
from several different organs
\cite{kuepper2016label,kallenbach2013immunohistochemistry}. However,
the raw spectra obtained from FTIR imaging experiments inherently
suffer from the Mie scattering effect
\cite{mohlenhoff2005mie,miljkovic2012line}, which affects the measured
absorption spectra and complicates the data
analysis \cite{bassan2009dispersion}. 

The underlying scattering effect is observable when applying FTIR
imaging to biological samples. Here, cells, nuclei or other cellular
components within a certain size range
\cite{mohlenhoff2005mie,miljkovic2012line} lead to a Mie scattering
effect. This model led to the development of first correction
procedures \cite{kohler2008estimating} based on the \emph{extended
  multiplicative signal correction} algorithm
\cite{martens1991extended}. This approach was extended by the authors
of \cite{bassan2010resonant}, who introduced an iterative correction
procedure for resonant Mie scattering (RMieS).  While this approach
takes into account scattering only, it has recently been further
improved upon by approximating the complete Mie extinction through
complex valued refractive indices of the scatterers
\cite{solheim2019open}. In short, an FTIR pixel spectrum observed in
an hyperspectral microscopic image is a mixture of Mie scattering and
resonant absorption factors, which has led to the development of
correspondingly complex computational correction procedures.

While the very recent ME-EMSC approach \cite{solheim2019open} promises
great improvement over the less elaborate scattering model of the
RMieS approach, our contribution is focused on the latter approach
\cite{bassan2010resonant}, which has been popular in a large range of
studies
\cite{kallenbach2013immunohistochemistry,kuepper2016label,witzke2019integrated}. Throughout
this manuscript, we will refer to the approach from
\cite{bassan2010resonant} as \emph{RMieS correction}. This approach
employs a reference spectrum, which represents an idealized baseline
of a scattering-free infrared spectrum. This spectrum is used
iteratively to approximate the measured, distorted spectra to the pure
absorbance spectrum using the \emph{extended multiplicative signal
  correction} \cite{bassan2010resonant}. Because of its iterative
nature there is a strong trade-off of between time and accuracy to
reach satisfactory results, making it computationally
expensive.\\

In a recent contribution, we demonstrated that in the presence of
sufficient data for training, deep neural networks may circumvent
RMieS correction algorithm \cite{raulf2020deep}. This approach is
based on an approach introduced in the context of \emph{representation
  learning} \cite{bengio2013representation}, specifically by employing
the approach introduced in \cite{rifai2011contractive} to perform
unsupervised pre-training followed by supervised fine-tuning to
classify pixel spectra into a discrete set of classes, i.e., tissue
components.

While the neural network introduced in \cite{raulf2020deep} involves
training data obtained from RMieS corrected spectra and thus involves
RMieS correction in an implicite manner, the model possesses no
explicit knowledge of Mie scattering. Yet, it has been hypothesized in
\cite{raulf2020deep} that, due to the strong generalization capability
of the network, it may have learned to disentangle the raw spectra
into an abstract representation that separates scattering from the
molecular spectrum. Our present contribution further investigates this
hypothesis by explicitly training the network to approximate
the complex function computed by the RMieS correction procedure. The
rationale behind our present study is roughly as follows: We replace
the final layer of a pretrained classifying neural network by a
regression layer to learn RMieS correction -- if supervised finetuning
of the pretrained regression network successfully learns RMieS
corresction, this provides evidence about the disentanglement in the
classifying network, namely that the pretraining helps to disentangle
those variances that are due to resonant Mie scattering.

\section{Methods}

\subsection{Dataset}

For our study, we used data sets from \cite{raulf2020deep} and
\cite{kallenbach2013immunohistochemistry} that we briefly recapitulate
for the sake of completeness. All samples were recruited from
thin-sections of colon cancer associated tissue samples. Two types of
tissue were used. Our first set of samples was recruited from formalin
fixed parrafin embedded (FFPE) histopathological samples, and the
second set from fresh frozen (FF) tissue. The samples were further
subdivided into one dataset $\mathrm{FFPE}_{\mathrm{pt}}$ for
pretraining (see Section \ref{sect:approach}), and one dataset
$\mathrm{FFPE}_{\mathrm{ft}}$ for supervised training (finetuning) the
regression-model. The tissue microarray data from \cite{raulf2020deep}
were used for training using an identical subdivision into training
and validation data as described in \cite{raulf2020deep}. Whole-slide
images from \cite{kallenbach2013immunohistochemistry} were used as
independent test sets.

\section{Approach} \label{sect:approach}

\begin{figure*}
  \centering
  \includegraphics[width=.7\textwidth]{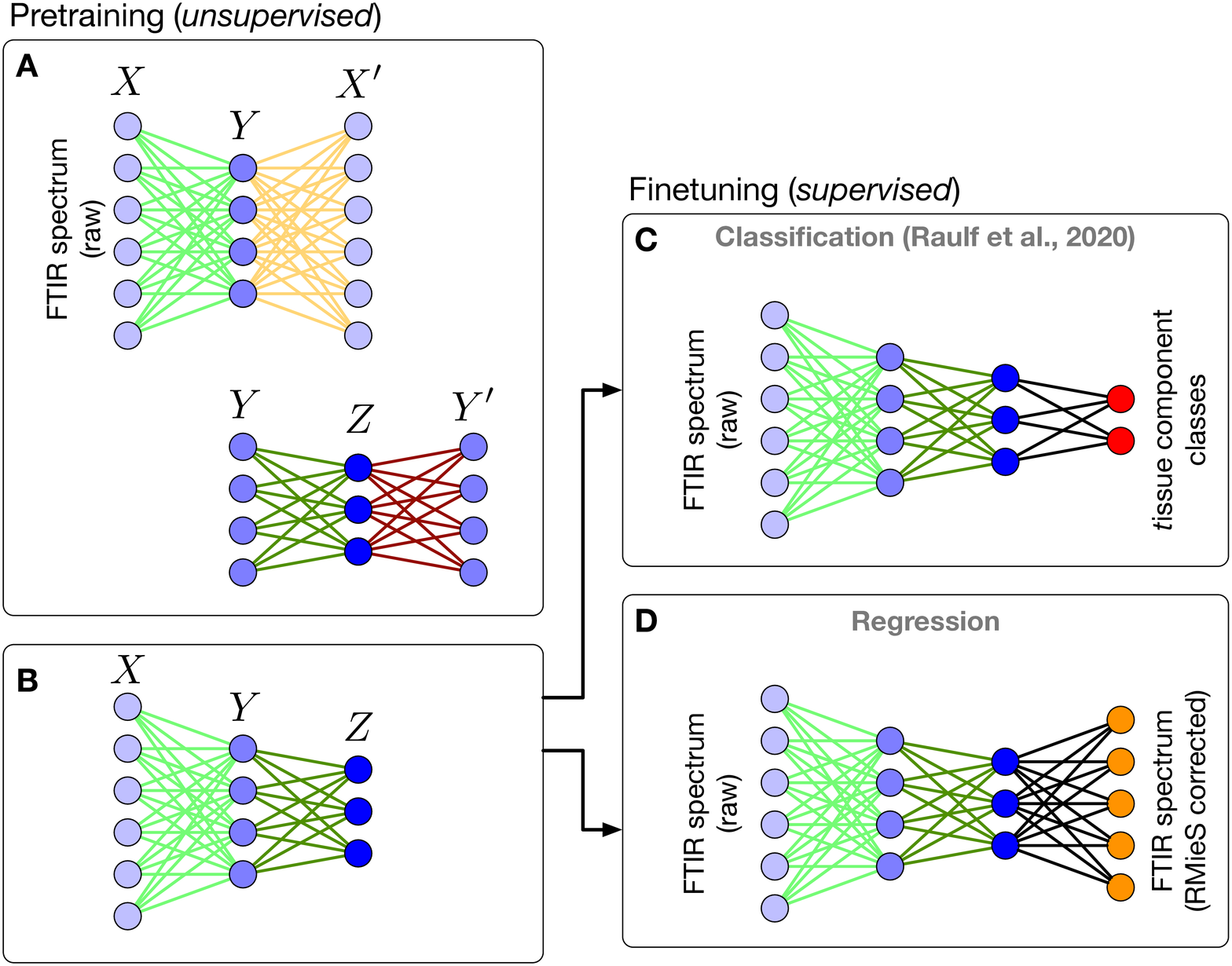}
  \caption{Overview of our approach to train a regression network
    (panel \textbf{D}) that approximates RMieS correction based on
    unsupervised pretraining through stacked autoencoders (panels
    \textbf{A} and \textbf{B}). The approach is similar to the tissue
    component classifier proposed in \cite{raulf2020deep} (panel
    \textbf{C}) Each output neuron of the regression network
    (indicated in orange in panel \textbf{D}) learns regression of one
    specific wavenumber of the RMieS corrected spectrum.}
\label{fig:overview}
\end{figure*}

Our general approach is to extend the stacked autoencoder based
network topology and training procedure from \cite{raulf2020deep} for
classifying infrared pixel spectra to obtain a neural network that
approximates the RMieS correction procedure from
\cite{bassan2010rmies}, as illustrated in Figure
\ref{fig:overview}. In fact, we used the RMieS correction
implementation described in \cite{bassan2010rmies} to produce training
and validation data; to introduce some essential notation, we denote
an RMieS corrected spectrum $y=R(x)$, where $R$ denotes the RMieS
correction procedure and $x$ an uncorrected raw spectrum from one of
the data sets.

Specifically, we use the paradigm of unsupervised pretraining as
established in \cite{hinton2006reducing,rifai2011contractive}, where
an unsupervised pretraining on unlabeled data is used to give the
initial mode for the used weight matrices in further training
stages. While in \cite{raulf2020deep}, these pretrained models
underwent supervised finetuning to train a classifier network, this
present contribution deals with a regression network aiming to
approximate the RMieS correction function rather than aiming to
classify pixel spectra. In other words, we deal with a neural network
whose output layer represents Mie-corrected infrared spectra. To this
end, we replace the transfer function of the output layer from a
softmax function commonly used for classifying networks to a linear
activation function suiting the requirements of a regression
model. All regression models are based on an unsupervised Contractive
Stacked Autoencoder (CSAE) \cite{rifai2011contractive} which was
trained only on the $\mathrm{FFPE}_{pt}$ dataset. Throughout the
paper, we will use $\theta$ to denote the parameters obtained from
supervised finetuning, and $y=S_\theta(x)$ the network with parameters
$\theta$ applied to input spectrum $x$, i.e., the approximation of the
corrected spectrum of $x$. During training, we used root mean square
error as loss function.

\subsubsection*{Validation Measures}

We validate our trained model $\theta$ on each of the validation data
sets $F$ at three levels. At the first level, we investigate the root
mean square error $RMSE_\theta=\sum_{x\in F}\|R(x)-S_\theta(x)\|$. On a
second level of validation, we used an existing random forest based
classifier $C$ from a previous study \cite{kuepper2016label} that
classifies a Mie corrected spectrum $y$ into one out of nineteen
different tissue component classes $C(y)$, and compared the output
classes of the ground truth $C(R(x))$ with the classification obtained
from an approximated correction, i.e., $C(S_\theta(X))$. We will refer
to the classifier $C$ as a \emph{downstream} model and thus refer to
this validation approach as \emph{downstream validation}.

On a third level of validation, we assess unvertainty of the trained
regression model based on the Bayesian dropout approach proposed by
Gal \textit{et al} \cite{gal2016dropout}, which systematically
integrates the concept of \emph{dropout layers} (i.e., the randomized
dropping of neurons in specific layers) into an approximation of a
Gaussian process.  The statistical processes can be introduced into
trained neural networks by using the usual dropout
\cite{srivastava2014dropout} not only as a tool to prevent overfitting
on the training dataset but also during the test phase to randomly
exclude $50\%$ of neurons at test time. By excluding neurons at test
time, one obtains a Bernoulli distribution over all different models
of the trained network, which approximates the variational inference
and finally approximates the deep Gaussian process. The latter step
yields a tool to interpret deep neural networks as models by
considering the prediction itself, the mean of the prediction and the
variance of this process.

The RMieS correction procedure is also highly time sensitive, which led
us to validate the running time difference between the RMieS correction
reference implentation and its neural network approximator. As an
iterative approach that needs to be applied to each individual pixel
spectrum in an infrared microscopic image, practical running times can
amount to hours when dealing with whole slide images that comprise
tens of millions of pixel spectra
\cite{kallenbach2013immunohistochemistry}, where several iterations of
the RMieS correction procedure may be required to achieve high quality
corrected spectra. 
At the same time, it is not straightforward
to implement the RMieS correction algorithm in a way that the
parallelization capability of graphics hardware can be fully exploited
\cite{bassan2010rmies}. Here, the potential promise of an approximator
network is a large increase in processing speed, since common neural
network frameworks can inherently and fully exploit parallelization
capability.

\paragraph*{Implementation}\ We utilized two implementations of the
RMieS correction provided by the authors of
\cite{bassan2010resonant}. Henceforth, we will refer to these
implementations as \emph{EMSC V2} and \emph{EMSC V5},
respectively. The network $S_\theta$ was trained using raw spectra as
input and \emph{EMSC V2} corrected spectra as target output for
regression learning. The \emph{EMSC V5} implementation was used as a
reference. All neural networks were implemented using the Theano
framework, as described in \cite{raulf2020deep}.

\section{Results}

\begin{figure}[!h]
\center
\includegraphics[width=\textwidth]{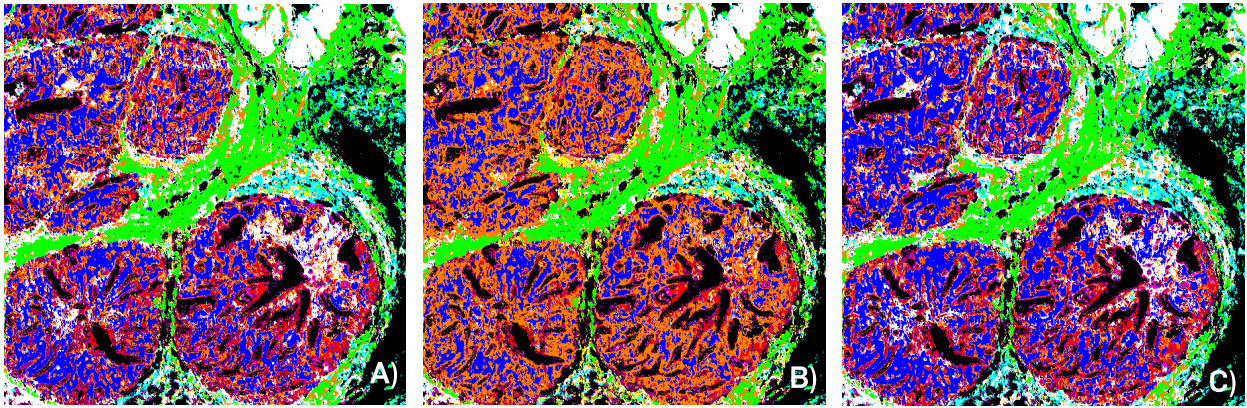}
\caption{Panel \textbf{A} displays classification results of the
  random forest classifier from \cite{kuepper2016label} applied to the
  FTIR spectra corrected with the \emph{EMSC V2} implementation of the
  RMieS correction. Panel \textbf{B} displays classification results
  obtained from the same classifier, but spectra corrected using the
    \emph{EMSC V5} implementation of the FTIR spectra \cite{kallenbach2013immunohistochemistry}. Panel \textbf{C}
  displays spectra corrected by the regression network $S_\theta$ that
  was trained to approximate the correction as implemented in
  \emph{EMSC V2}.} \label{fig:downstream-ffpe}
\end{figure}

\begin{figure*}[!h]
\center
\includegraphics[scale=0.2]{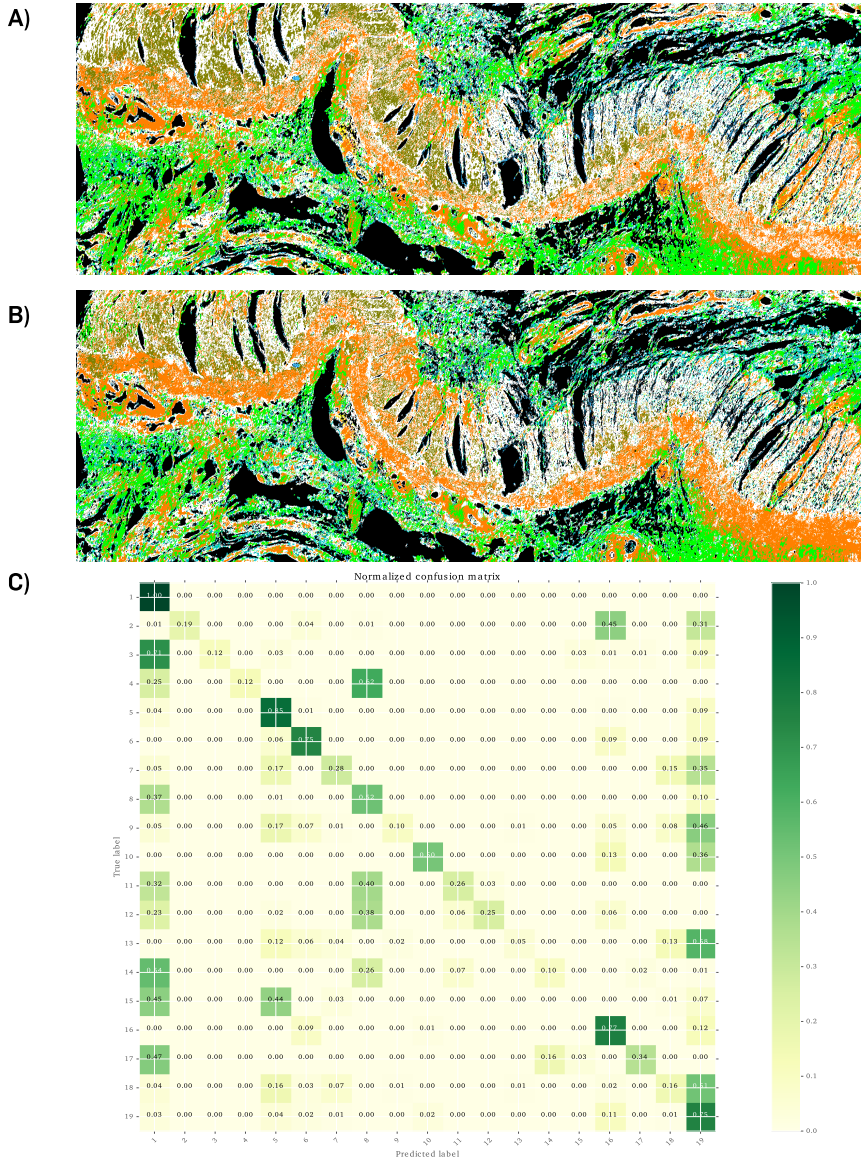}
\caption{Panel \textbf{A} displays classification results of the
  random forest classifier from \cite{kuepper2016label} applied to the
  FTIR spectra corrected with the \emph{EMSC V2} implementation of the
  RMieS correction. Panel \textbf{B} displays spectra corrected by the
  regression network $S_\theta$ that was trained to approximate the
  correction as implemented in \emph{EMSC V2}. A normalized confusion
  matrix is displayed in panel \textbf{C}.}
  \label{fig:downstream-whole-slide}
\end{figure*}

\paragraph*{Downstream validation.}\ Figure \ref{fig:downstream-ffpe}
shows the comparison of the validation dataset for the FFPE data using
the random forest introduced in \cite{kuepper2016label} as downstream
classifier that classifies RMieS corrected spectra into one out of 19
different tissue components. Compared to the ground truth
segmentation obtained from $C(R(x))$ for each pixel spectrum $x$, the
approximation based classification constituted by $C(S_\theta(x))$
achieves an accuracy of $78\%$ across all pixels in the whole-slide
image displayed in Figure \ref{fig:downstream-whole-slide}.

\paragraph*{Running time.}\ To assess running times, we performed
correction of a validation data set of size $600*600$ spectra. The
time that has been recorded was averaged over 10 different runs each
and are summarized in the Table \ref{tab:running-times}.

\begin{table}
  \centering
  \begin{tabular}{|c|c|c|}
    \hline 
    Model & EMSC & NN \\ 
    \hline 
    Time for val.-set & 64.99 sec &  10.65 sec \\ 
    \hline 
    Time per spectrum & 118.23 $\mu$sec & 28.96 $\mu$sec  \\ 
    \hline 
  \end{tabular} 
  \caption{Running times obtained from RMieS correction reference implementation
    (\emph{EMSC}) and the approximation by neural network (\emph{NN}) for a data set size of 360000 spectra. Recorded times were averaged over 10 runs each.}
\label{tab:running-times}
\end{table}

\paragraph*{Characterization of approximation capabilities.}\ 

\begin{figure*}[!h]
\center
\includegraphics[width=.7\textwidth]{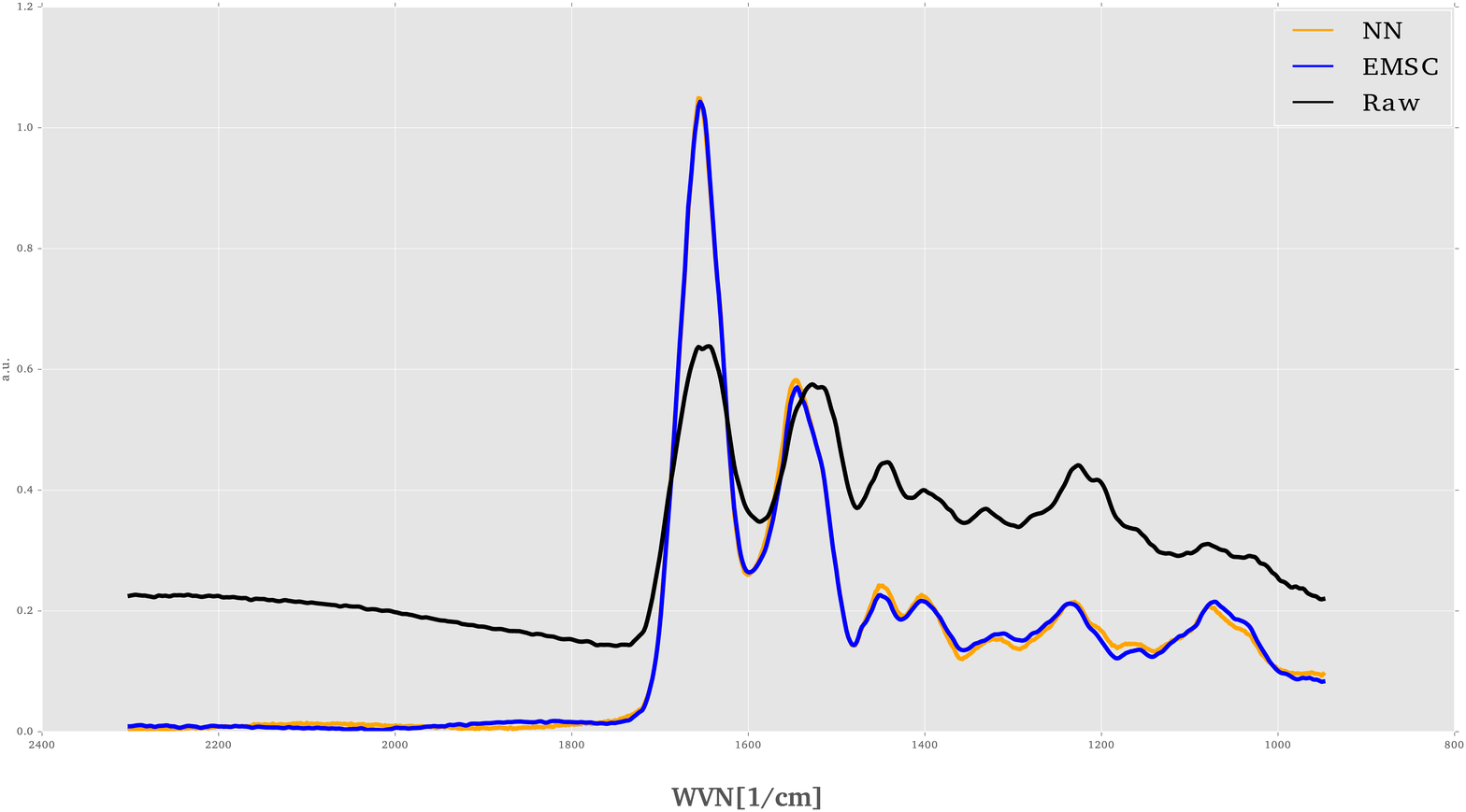}
\caption{Example of an FTIR spectrum from FFPE tissue, shown as raw
  spectrum (black), corrected by the RMieS correction algorithm from
  \cite{bassan2010resonant} (blue) and corrected by the neural network
  $S_\theta$ that approximates the RMieS correction.
  EMSC.} \label{fig:result-example}
\end{figure*}

\begin{figure*}[!h]
\center
\includegraphics[width=.5\textwidth]{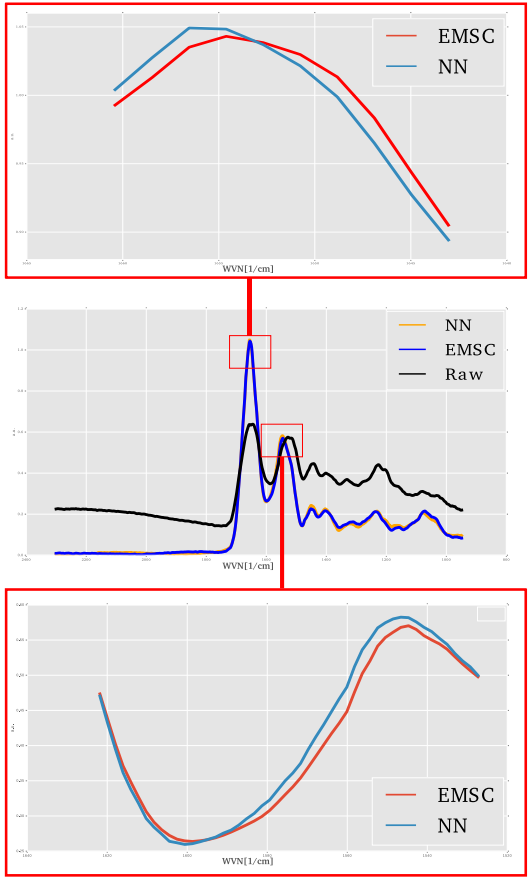}
\caption{Comparison of band shift by reference implementation of RMieS
  correction (\emph{EMSC}) and the approximating neural network
  (\emph{NN}).} \label{fig:band-shift}
\end{figure*}

\begin{figure*}[!h]
\center
\includegraphics[scale=0.5]{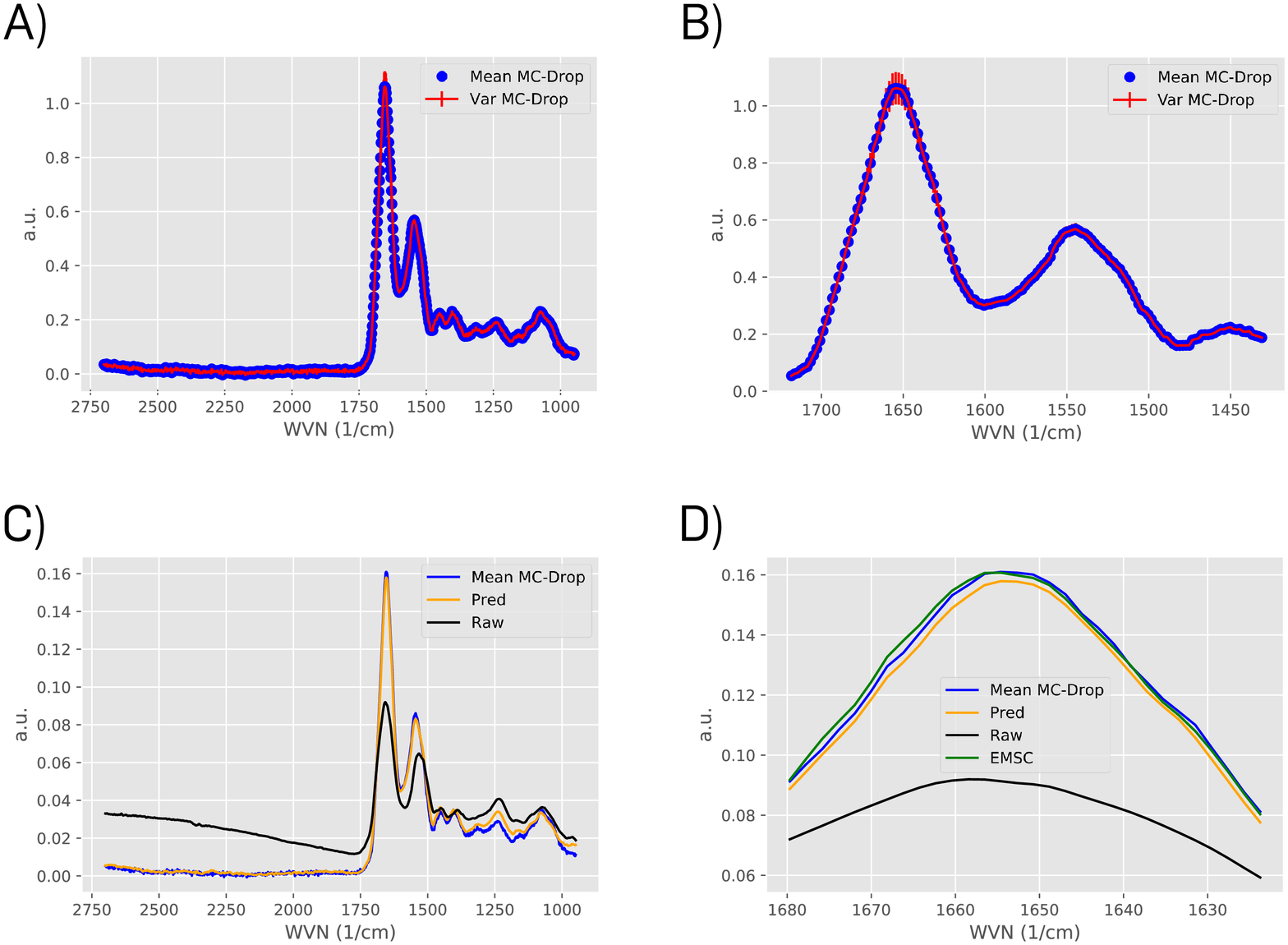}
\caption{\emph{Panel \textbf{A}}: Mean spectrum obtained by neural
  network $S_\theta$ including confidence interval obtained by MC
  dropout; \emph{\textbf{B}}: Same as \emph{\textbf{A}}, focusing on
  the Amide bands of the spectrum and demonstrating the relatively low
  confidence in the regression around the amide I peak; \emph{Panel
    \textbf{C}}: Comparison of mean spectra; Panel \emph{\textbf{D}}:
  Same as \emph{\textbf{C}}, focusing on the Amide bands of the
  spectrum.} \label{fig:dropout}
\end{figure*}

As indicated in Figure \ref{fig:result-example} and panel \textbf{C}
of Figure \ref{fig:band-shift}, the corrected spectra obtained from
network $S_\theta$ approximate the RMieS correction function with only
little error. However, the deviation around the amide I peak around
1650 cm$^{-1}$ is remarkably high. In fact, detailed inspection
(Figure \ref{fig:band-shift}) indicates a band shift between the RMieS
corrected ground truth spectrum and the neural network
approximation. To further assess this band shift, we performed
Bayesian dropout validation, which yields a confidence interval at
each wavenumber, as displayed in Figure \ref{fig:dropout}. The
confidence intervals are strikingly large around the amide I peak. In
other words, the band shift coincides with a low-confidence region of
the network.

\section{Conclusion}

Our results clearly demonstrate that that the RMieS correction for
infrared spectra can be approximated by a neural network that produces
practically useful corrected spectra, while using only a fraction of
the computation time. Beyond the immediate and practically highly
relevant benefit in terms of computational speedup, our results also
contribute to the understanding and interpreting of what deep neural
network models have learned during supervised training. In fact, in
\cite{raulf2020deep} it was hypothesised that autoencoder-based
pretraining for a \emph{classifying} neural network may have learned
to disentangle raw infrared pixel spectra in a manner such that the
variance due to resonant Mie scattering has been separated from the
variance that is due to vibrations at the molecular level. The fact
that the same pretrained stacked autoencoder allows to compute
corrected spectra adds further support to this hypothesis.


In general, it is important to keep in mind the inherent limitations
of approximations obtained from deep neural networks as the one we
have introduced here. In fact, the network function $S_\theta$ we
obtain is a very local approximation of the RMieS correction function
$R$ in the sense that it works primarily for input spectra that
sufficently resemble the training data. In other works, as long as a
raw spectrum $x$ is obtained from FFPE samples of colon tissue,
applied to similar substrate and spectroscopically measured in a
similar manner, then $S_\theta(x)$ will produce spectra that will
reliably resemble $R(x)$. It is a highly relevant question for future
research to train networks that work reliably on a broader set of
inputs, e.g. across tissue from different organs and being either FFPE
or fresh-frozen as well as potentially being prepared on different
substrate material.

Even with the limited generalization guarantee resulting from
relatively limited training data, the computational speedup
constitutes a factor that makes our results promising from a practical
perspective, since the high demand of computation time can easily
become a road block in many pracitcal setting, when e.g. dealing with
whole slide images.




\bibliographystyle{abbrv}

\end{document}